\documentclass[letterpaper, 10 pt, conference]{ieeeconf}  %

\IEEEoverridecommandlockouts                              %

\usepackage{graphics} %
\usepackage{epsfig} %
\usepackage{eso-pic} %
\usepackage{times} %
\usepackage{amsmath} %
\usepackage{amssymb}  %
\usepackage{multirow}
\usepackage{tikz}
\usetikzlibrary{spy}
\usepackage{pgfplots}
\usepackage{pgfplotstable}
\usepgfplotslibrary{groupplots}
\usepackage{booktabs}
\usepackage{subcaption}
\usepackage{makecell}
\usepackage{tabularx}
\usepackage{xcolor}
\usepackage{svg}
\usepackage[pagebackref=true,breaklinks=true,bookmarks=false,colorlinks=true]{hyperref} %
\usepackage[table]{xcolor}

\pgfplotsset{compat=1.18}

\definecolor{barrier}{RGB}{255,120,50}
\definecolor{bicycle}{RGB}{255,192,203}
\definecolor{bus}{RGB}{255,255,0}
\definecolor{car}{RGB}{0,150,245}
\definecolor{construction_vehicle}{RGB}{0,255,255}
\definecolor{motorcycle}{RGB}{200,180,0}
\definecolor{pedestrian}{RGB}{255,0,0}
\definecolor{traffic_cone}{RGB}{255,240,150}
\definecolor{trailer}{RGB}{135,60,0}
\definecolor{truck}{RGB}{160,32,240}
\definecolor{driveable_surface}{RGB}{255,0,255}
\definecolor{other_flat}{RGB}{139,137,137}
\definecolor{sidewalk}{RGB}{75,0,75}
\definecolor{terrain}{RGB}{150,240,80}
\definecolor{manmade}{RGB}{230,230,250}
\definecolor{vegetation}{RGB}{0,175,0}

\definecolor{m_green_border}{HTML}{82B366}
\definecolor{m_darkgreen}{HTML}{d5e9d5}
\definecolor{m_green}{HTML}{d5e9d5}
\definecolor{m_orange}{HTML}{fff2cd}
\definecolor{m_orange_border}{RGB}{150,114,164}
\definecolor{m_red}{HTML}{fccecd}
\definecolor{m_violet}{HTML}{e2d6e8}
\definecolor{m_blue}{HTML}{d8e9fc}
\definecolor{m_queries}{RGB}{195,212,217}
\definecolor{m_queries_border}{RGB}{103,103,104}
\definecolor{m_red}{RGB}{248,206,204}
\definecolor{m_red_border}{RGB}{204,29,12}
\definecolor{m_z_dens}{RGB}{255,223,209}
\definecolor{m_z_dens_border}{RGB}{250,139,139}
\definecolor{m_blue_border}{RGB}{107,141,190}
\definecolor{m_yellow}{RGB}{255,242,205}
\definecolor{m_yellow_border}{RGB}{213,182,82}
\definecolor{m_lidar}{RGB}{255,230,128}
\definecolor{m_lidar_border}{RGB}{103,103,104}
\definecolor{m_camera}{RGB}{200,196,183}
\definecolor{m_camera_border}{RGB}{103,103,104}
\definecolor{m_attn}{RGB}{135,222,170}
\definecolor{m_attn_border}{RGB}{103,103,104}
\definecolor{m_gray}{RGB}{245,245,245}
\definecolor{m_gray_border}{RGB}{102,102,102}
\newcommand{\colorsquare}[1]{\tikz{\path[draw=#1_border,fill=#1, thick, rounded corners=0.6pt] (0,0) rectangle (6pt,6pt);}}

\newcommand{\PAR}[1]{\vskip4pt \noindent {\bf #1~}}

\renewcommand*\backref[1]{\ifx#1\relax \else #1 \fi}

\newcolumntype{K}[1]{>{\centering\arraybackslash}p{#1}}

\newif\ifmynotes
\mynotestrue

\newcommand{\cb}{\cellcolor{blue!10}}

\newcommand\copyrighttext{%
    \footnotesize \copyright{ }2026 IEEE. Personal use of this material is permitted. Permission from IEEE must be obtained for all other uses, in any current or future media, including reprinting/republishing this material for advertising or promotional purposes, creating new collective works, for resale or redistribution to servers or lists, or reuse of any copyrighted component of this work in other works.}
\newcommand\copyrightnotice{%
    \AddToShipoutPicture*{%
        \put(\LenToUnit{.5\paperwidth-.5\textwidth}, \LenToUnit{30pt}){%
            \parbox{\textwidth}{\copyrighttext}%
        }%
    }%
}

\title{\LARGE \bf
Towards Trustworthy and Explainable AI for Perception Models: \\ From Concept to Prototype Vehicle Deployment
}

\author{Till Beemelmanns, Shayan Sharifi, Manas Mehrotra, Ayushman Choudhuri, Lutz Eckstein%
\thanks{This work has received funding from the European Union’s Horizon Europe Research and Innovation Programme under Grant Agreement No. 101076754 - AIthena project. The authors are with the Institute for Automotive Engineering, RWTH Aachen University, 52074 Aachen, Germany}%
}

\begin{document}

\maketitle
\thispagestyle{empty}
\pagestyle{empty}

\copyrightnotice

\begin{abstract}
Deep Neural Networks have become the dominant solution for Autonomous Driving perception, but their opacity conflicts with emerging Trustworthy AI guidelines and complicates safety assurance, debugging, and human oversight.
While theoretical frameworks for safe and Explainable AI (XAI) exist, concrete implementations of Trustworthy AI for 3D scene understanding remain scarce.
We address this gap by proposing a \emph{Trustworthy AI} perception module that is remarkably robust, integrates faithful explainability, and calibrated uncertainty estimates.
Building on a transformer-based detector, we derive explanation from the attention mechanism at inference time and validate their faithfulness using perturbation-based consistency tests.
We further integrate an uncertainty estimation and calibration module, and apply robustness-enhancing training methods.
Experiments show faithful saliency behavior, improved robustness, and well-calibrated uncertainty estimates.
Finally, we deploy these Trustworthy AI elements in a prototype vehicle and provide an XAI Interface that visualizes documentation artifacts, model uncertainty state, and saliency maps, demonstrating the feasibility of trustworthy perception monitoring in real time.
Supplementary materials are available at \href{https://tillbeemelmanns.github.io/trustworthy_ai/}{https://tillbeemelmanns.github.io/trustworthy\_ai/}.
\end{abstract}

\section{Introduction}
Autonomous Driving (AD) perception has achieved impressive accuracy by leveraging Deep Neural Networks (DNNs) for multi-sensor 3D scene understanding~\cite{yan2023cross,xie2023sparsefusion}.
However, these gains come with a core mismatch: AI perception models are inherently opaque, yet safety-critical deployment demands evidence for safety assurance when the system may fail, when it is uncertain, and how a human operator should interpret its outputs.
This is in tension with Trustworthy AI guidelines, as emphasized by the EU Ethics Guidelines~\cite{eu}, the NIST AI Risk Management Framework~\cite{nist}, and with safety standards such as ISO~26262 and ISO~21448 (SOTIF), which were not designed for black-box systems~\cite{iso26262_2018,iso21448_2022}.
Deployed AI perception systems therefore need mechanisms for transparency that support safety assurance, debugging, and human oversight, including interfaces that surface such information to authorities, operators, and developers.
Recent work on AI safety in AD emphasizes the need for \emph{Trustworthy AI} combining explainability, robustness with a human-machine interface~\cite{kuznietsov2024explainable}.
Trustworthy AI is also characterized by the use of uncertainty; we expect an AI to quantify its true level of (un)certainty, providing a reliable signal for downstream modules or other agents to handle unreliable output~\cite{ts,trust,dece}.
While theoretical Trustworthy AI frameworks for AD propose modular architectures for interpreting and monitoring AI modules, concrete implementations remain scarce~\cite{kuznietsov2024explainable}.
Specifically, existing Explainable AI (XAI) methods predominantly focus on the 2D image domain, leaving LiDAR and multi-view camera 3D scene understanding methods largely unexplored~\cite{Ribeiro2016LIME,Selvaraju2017GradCAM,Lundberg2017SHAP}.
Furthermore, most research operates in offline settings, lacking quantitative validation of the explanations or integration into real-world demonstrators.

\begin{figure}[t]
\centering
\includegraphics{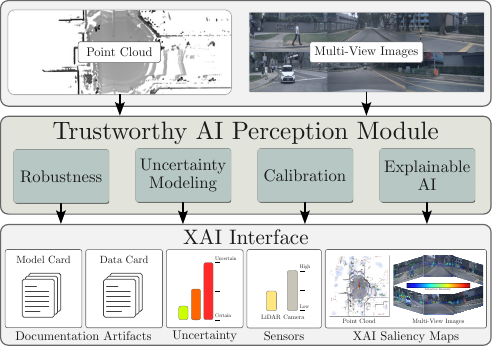}
\caption{
\textbf{Proposed Trustworthy AI Approach.}
We propose a multi-modal perception module that integrates Trustworthy AI components: robust training, calibrated uncertainty quantification, and explainability.
An XAI Interface enables transparent monitoring through documentation, visualized uncertainty state, sensor usage, and saliency maps.
}
\label{fig:teaser}
\end{figure}

In this work, we bridge the gap between abstract safety requirements and practical deployment.
We propose a Trustworthy AI perception module, illustrated in Fig.~\ref{fig:teaser}, that extends LiDAR-camera 3D object detection with XAI, Uncertainty Quantification (UQ), and robustness-enhancing training.
Unlike traditional XAI methods that are computationally expensive, we leverage the integrated attention mechanisms of the transformer-based module to generate real-time explanations.
To ensure these explanations are not merely visual aids but faithful representations of model behavior, we subject them to perturbation-based consistency testing~\cite{Fong2017MeaningfulPerturbation}.
Simultaneously, we integrate an efficient UQ module that provides uncertainty estimates, enabling downstream modules to identify unreliable predictions.
We move beyond simulation and dataset-only evaluation by instantiating these concepts in a real-world research vehicle focusing on a LiDAR-based perception module.
An \emph{XAI Interface} visualizes the model's explainability features and uncertainty state to a human operator in real time, complemented by documentation artifacts.
This demonstrates how Trustworthy AI components can support debugging, auditability, and operational safety monitoring.
Our main contributions are as follows:
\textbf{(1)} We introduce a Trustworthy AI perception component with strong robustness, calibrated uncertainty estimates, and faithful explainability features;
\textbf{(2)} We address the lack of quantitative evaluation in 3D scene understanding by conducting perturbation tests to verify XAI feature faithfulness and evaluate 3D uncertainty calibration;
\textbf{(3)} We present an integration of these Trustworthy AI components into a real-world demonstrator vehicle, proving the feasibility of our approach in practice.

\section{Related Work}

\PAR{Trustworthy AI.}
General frameworks, such as the EU Ethics Guidelines~\cite{eu} and the NIST AI Risk Management Framework~\cite{nist}, define Trustworthy AI through principles of technical robustness, data governance, accountability, and transparency.
In the context of AD, Kuznietsov et al. synthesize these requirements into three domains, \emph{data} (governance and diversity), \emph{model} (robustness and explainability), and \emph{agency} (human oversight)~\cite{kuznietsov2024explainable}.
Based on five main XAI paradigms, Kuznietsov et al. propose the SafeX framework, which provides a conceptual, modular architecture for integrating XAI into the AD stack to support safety and trustworthiness, via a combination of \emph{interpretable design}, \emph{monitoring}, and \emph{auxiliary explanations}.
In parallel, recent methodologies advocate for the use of structured reporting documents, such as \emph{Model Cards} and \emph{Data Cards}~\cite{canas2024methodology}, to address transparency requirements.
Model Cards provide a standardized disclosure of a model's intended use, architectural details, and performance across diverse conditions, while Data Cards document the composition and potential biases of the underlying training sets.
These cards have been tailored to map technical specifications directly to ethical and regulatory obligations, such as the EU AI Act~\cite{canas2024methodology}.

In contrast to purely conceptual frameworks, our work instantiates key Trustworthy AI components for a 3D scene understanding module.
We bridge the gap between high-level principles and implementation by coupling Uncertainty Quantification (UQ) and Explainable AI (XAI) with a dedicated interface.
Furthermore, we enhance system auditability by integrating comprehensive documentation through Model and Data Cards, providing a transparent record of the system's design, capabilities, and limitations.

\PAR{Explainability.}
XAI aims to make machine learning models transparent and interpretable for human users~\cite{Molnar2022InterpretableML}.
DNNs often act as black boxes, providing limited insight into their internal reasoning, which complicates safety assessment and debugging~\cite{Molnar2022InterpretableML}.
XAI helps to reveal how models arrive at decisions, thereby improving user confidence and supporting verification workflows~\cite{Amershi2019Guidelines}.
A central distinction in XAI lies between global explanations, which characterize a model’s overall behavior, and local explanations, which focus on individual predictions~\cite{Molnar2022InterpretableML}.
Popular local explanation approaches include LIME~\cite{Ribeiro2016LIME} and SHAP~\cite{Lundberg2017SHAP}: LIME approximates a model locally using interpretable surrogate functions, while SHAP provides feature attributions grounded in game theory.
In computer vision, XAI methods frequently appear as saliency maps that highlight influential image regions. %
Gradient-based saliency maps estimate pixel importance by differentiating a target score with respect to the input image~\cite{Simonyan2014Saliency}, whereas Grad-CAM~\cite{Selvaraju2017GradCAM} derives a coarse localization heatmap by backpropagating gradients to feature maps and using them to weight the corresponding activations~\cite{Selvaraju2017GradCAM}.
Perturbation-based methods~\cite{Fong2017MeaningfulPerturbation}, such as RISE~\cite{rise} and D-RISE~\cite{Petsiuk2021drise}, provide causal insight by masking parts of the input and measuring the resulting effect on predictions.

This principle has also been extended to LiDAR-based 3D detectors, for example, OccAM’s Laser removes subsets of point clouds to analyze feature importance~\cite{Schneider2022OccAM}.
Attention-based explanations leverage attention weights as a token-level attribution signal, providing a structured view of which inputs are emphasized during inference~\cite{Vaswani2017Attention,beemelmanns2023explainable}.
However, prior studies show that raw attention weights do not always reflect true causal influence, motivating the use of sanity checks and perturbation validation~\cite{Jain2019AttentionNotExplanation}.
In this work, we validate the faithfulness of attention-derived explanations via perturbation tests and use attention as our primary explanation signal, since methods such as LIME/SHAP~\cite{Ribeiro2016LIME,Lundberg2017SHAP} and (D-)RISE~\cite{rise,Petsiuk2021drise} require many forward passes per sample, and Grad-CAM~\cite{Selvaraju2017GradCAM} requires backpropagation, making them impractical for real-time deployment.

\PAR{Robustness.}
AD perception must remain reliable under distribution shifts and input corruptions, since a real-world deployment experiences calibration errors, temporal misalignment, and heterogeneous sampling rates~\cite{beemelmanns2024multicorrupt}. %
Sensor data can be partially missing or corrupted due to hardware faults, packet loss, and environmental effects such as rain, fog, snow, glare, or motion blur.
These factors can cause performance degradation in 3D understanding pipelines, motivating dedicated robustness benchmarks~\cite{beemelmanns2024multicorrupt,kong2023robo3d,xie2025benchmarking}.

One evaluation direction uses real-world adverse-condition datasets, such as Seeing Through Fog~\cite{bijelic2020seeingfog} or Canadian Adverse Driving Conditions~\cite{Pitropov_2020cadc}. %
While these datasets provide realistic weather and visibility effects, their coverage is constrained by the high cost of data acquisition and manual annotation, and they often differ substantially in sensor setup and scene statistics from popular perception datasets, such as nuScenes~\cite{caesar2020nuscenes} and Waymo~\cite{sun2020scalability}, which limits the utility of such datasets as robustness benchmarks.

A complementary line of work constructs corruptions on top of established datasets to isolate specific failure modes while preserving the underlying sensor setup. %
RoboBEV~\cite{xie2025benchmarking} targets \emph{multi-view camera} perception and defines a robustness benchmark derived from nuScenes~\cite{caesar2020nuscenes}, featuring eight corruption types evaluated across three severity levels.
Robo3D~\cite{kong2023robo3d} analogously benchmarks \emph{LiDAR-only} perception, simulating eight corruption types to analyze the resilience of 3D detection and segmentation models.

Beyond modality-specific datasets, another line of work studies \emph{multi-modal} LiDAR-camera corruptions, such as missing camera inputs and temporal/spatial misalignment~\cite{Yu2022BenchmarkingTRALI,beemelmanns2024multicorrupt}.
MultiCorrupt~\cite{beemelmanns2024multicorrupt} introduces multi-modal corruptions into nuScenes, including adverse-weather effects alongside other cross-sensor failures, to evaluate robustness under a broader set of conditions.
Furthermore, MultiCorrupt analyzes and reveals best practices for the architectural design and training procedures of state-of-the-art multi-modal perception models.
We build upon the insights of MultiCorrupt, by using a robust multi-modal model architecture and by using a robustness-increasing training to further increase the trustworthiness of the perception module.

\PAR{Uncertainty Quantification.}
Approaches that estimate predictive uncertainty in DNNs are commonly grouped into: Bayesian approximations, ensemble methods, and parametric uncertainty models.
Bayesian approximations include Monte-Carlo Dropout (MCD)~\cite{mcd,postels2019mcd}, where dropout is kept active at inference to approximate sampling from a posterior over weights.
Deep Ensembles (DE)~\cite{de} evaluate multiple independently trained models simultaneously and the variance across their predictions is interpreted as predictive uncertainty.
Parametric models extend point estimates by predicting distribution parameters alongside the mean~\cite{Nix.1994,Heskes.1996}, enabling uncertainty estimation in a single pass. %

For \emph{2D detection}, prior studies model epistemic uncertainty via MCD~\cite{Harakeh.2020}, while aleatoric-based methods use separate variance prediction heads that directly regress the box-parameter variances~\cite{Le.2018, gaussianyolo}.
Direct regression of Gaussian distributions on the box parameters via minimizing the Kullback-Leibler~(KL) divergence was introduced by He et al.~\cite{He.23Sep18}, where the ground truth box parameters are modeled as Dirac delta distributions. %

For \emph{3D detection}, sampling-based methods apply MCD to LiDAR detectors~\cite{difeng2019mcd} or adopt the MIMO strategy~\cite{mimo,Pitropov.2022}.
These approaches produce multiple predictions per frame and require post-hoc clustering to obtain final detections and empirical uncertainties, which increases inference cost.
To avoid this overhead, single pass methods use either feature density estimators~\cite{occuq,beemelmanns2026q2u} or parametric approaches to predict uncertainties, e.g., variances on 3D box corners~\cite{Meyer.2019} or on box centroid and dimensions via heteroscedastic aleatoric modeling~\cite{difeng2019heteroscedastic}, or use Gaussian KL-based objectives~\cite{trust}.
For orientation, several works replace Gaussian assumptions with a von-Mises distribution to reflect angular periodicity~\cite{kld,coalign}.
In this work, we leverage a deployment-friendly KL-based uncertainty quantification approach to support trustworthy perception with point-wise predictive uncertainty.

\begin{figure*}[t]
\centering
\includegraphics[width=0.975\linewidth]{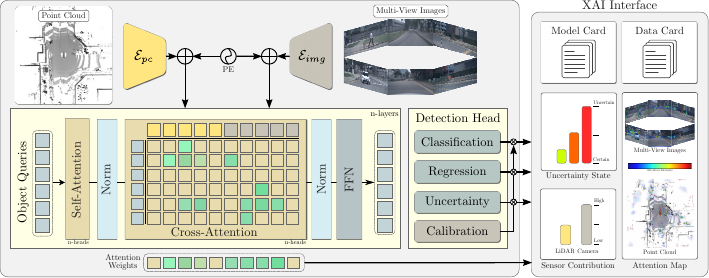}
\caption{
\textbf{Overview of the Trustworthy AI Approach.}
From LiDAR point cloud and multi-view camera images, we extract camera (\protect\colorsquare{m_camera}) and LiDAR tokens (\protect\colorsquare{m_lidar}) that interact with object queries (\protect\colorsquare{m_queries}) via Cross-Attention and produce calibrated uncertainty-aware 3D bounding boxes. 
Attention weights (\protect\colorsquare{m_attn}), derived sensor usage, and current model uncertainty state are used for visualization in the XAI Interface, along with supporting Model Cards and Data Cards.
}
\label{fig:method_overview}
\end{figure*}

\section{Method}
Guided by the principles of Trustworthy AI, we focus on three measurable components: (i) \emph{Explainability} for auditability and debugging, (ii) \emph{Uncertainty} to quantify predictive reliability, and (iii) \emph{Robustness} to maintain stable behavior under distribution shift.
Other Trustworthy AI elements (e.g., data privacy, fairness and societal considerations) depend largely on dataset governance and are therefore outside the scope of this work.
To introduce a Trustworthy AI perception module, we propose to leverage a transformer-based LiDAR-camera detector.
During runtime, we generate saliency maps and predict uncertainty estimates.
We propose an XAI Interface that presents the current model uncertainty state and input data contribution of the perception module, along with documentation artifacts.
Our approach is robust and explainable, and integrates these Trustworthy AI elements jointly into a real-world demonstrator.

\PAR{Method Overview.} %
We build on CMT~\cite{yan2023cross}, a transformer-based multi-modal 3D object detection architecture.
From multi-view cameras, we first extract features using a standard backbone $\mathcal{E}_{img}$, generate image tokens (Figure~\ref{fig:method_overview}, \colorsquare{m_camera}), and feed them into a transformer decoder.
Similarly, for the LiDAR input, we use a sparse point cloud encoder $\mathcal{E}_{pc}$ to obtain point cloud tokens (\colorsquare{m_lidar}).
For both image and LiDAR tokens, we apply positional encodings to preserve spatial information.
In the transformer decoder, a set of learnable object queries (\colorsquare{m_queries}) interact with the 3D sensor tokens and produce refined queries, which are then used to predict 3D object boxes along with classification and uncertainty scores. 
The cross-attention weights (\colorsquare{m_attn}) are extracted during inference and serve as an explanation of which 3D sensor tokens contributed to a particular detection.
To summarize the current state of this AI module, we propose an XAI Interface that visualizes attention maps, uncertainty scores, and a sensor contribution score that attributes each modality’s contribution to the predicted detections.
Additionally, we display Model Cards and Data Cards in the XAI Interface to document model version, training data, ethical considerations, and known limitations.

\subsection{Explainability}
\PAR{Cross-Attention Extraction.}
To obtain explanations aligned with the model’s fusion mechanism, we extract the cross-attention weights from the decoder during inference.
For each decoder layer $\ell\in\{1,\dots,L\}$, we retrieve the attention tensor
\begin{equation}
A^{(\ell)} \in \mathbb{R}^{H \times Q \times S},
\end{equation}
where $H$ is the number of heads, $Q$ the number of object queries processed by the decoder, and $S$ the number of source tokens (LiDAR tokens and image tokens concatenated).
Intuitively, $A^{(\ell)}[h,q,s]$ measures how strongly query $q$ attends to token $s$ under head $h$ at layer $\ell$.

\PAR{Query Selection.}
Raw attention includes many low-confidence queries that do not correspond to meaningful predictions.
We therefore reduce the final query set in two steps: (i) select the top-$K$ object queries, and (ii) filter them by a confidence threshold $\tau$.
Let $\mathcal{I}\subset\{1,\dots,Q\}$ denote the selected top-$K$ indices and $\mathcal{Q}=\{q\in\mathcal{I}\mid \mathrm{score}(q)\ge \tau\}$ the final valid set.
The filtered attention then reads,
\begin{equation}
A_{\text{final}}^{(\ell)} = A^{(\ell)}[:,\,\mathcal{Q},:]\ \in\ \mathbb{R}^{H \times Q_{\text{valid}} \times S},
\end{equation}
with $Q_{\text{valid}}=|\mathcal{Q}|$.

\PAR{Attention Fusion.}
Transformer attention is distributed across layers and heads; to obtain a single interpretable saliency signal, we aggregate attention across transformer layers and heads using \emph{Mean-Fusion};
\begin{equation}
A_{\text{layer}} = \frac{1}{L}\sum_{\ell=1}^{L} A_{\text{final}}^{(\ell)} \in \mathbb{R}^{H \times Q_{\text{valid}} \times S},
\end{equation}
\begin{equation}
A_{\text{head}} = \frac{1}{H}\sum_{h=1}^{H} A_{\text{layer}}[h,:,:] \in \mathbb{R}^{Q_{\text{valid}} \times S}.
\end{equation}
We then collapse the query dimension via max pooling to obtain a per-token importance score.
\PAR{Modality-specific Saliency Maps.}
Since the tokens are a concatenation of LiDAR and camera tokens $S=S_{pc} + S_{img}$, we split the attention into modality-specific components and reshape them back to their native grids.
For visualization and analysis, these maps are resized to the original BEV and image resolutions, enabling qualitative overlays and serving as the basis for targeted perturbation tests of explanation faithfulness.

\PAR{Sensor Contribution.}
To quantify the relative reliance on each sensor, we compute the fraction of total fused attention mass assigned to each of the seven sensors.
Let $\mathbf{A}\in\mathbb{R}^{1\times Q\times(S_{pc} + S_{img})}$ denote the fused saliency and let $\mathcal{F}_m$ be the index set of tokens belonging to modality $m$ (LiDAR or one camera).
The contribution for each sensor then reads,
\begin{equation}
C_m \;=\; \frac{\sum_{q=1}^{Q}\sum_{f\in\mathcal{F}_m} A_{1,q,f}}{\sum_{q=1}^{Q}\sum_{f=1}^{S_{pc} + S_{img}} A_{1,q,f}}.
\end{equation}
We visualize $C_m$ alongside detections and saliency maps to summarize per-sensor contribution.

\subsection{Robustness}

\PAR{Robust Training.}
To increase the model robustness, we evaluate a masked-modal training strategy~\cite{yan2023cross}.
During training, we randomly mask entire sensor modalities so that the detector is trained with (i) LiDAR-only, (ii) camera-only, and (iii) LiDAR-camera input.
This exposes the model to failure modes and discourages over-reliance on any specific sensor, while keeping the architecture unchanged.
We apply the same masked-modal training protocol to several multi-modal 3D detector baselines, such as SparseFusion~\cite{xie2023sparsefusion} and BEVFusion~\cite{liu2022bevfusion}, and evaluate their robustness on MultiCorrupt~\cite{beemelmanns2024multicorrupt}.
\PAR{Robustness Evaluation.}
We assess robustness as the mean Relative Resistance Ability, i.e., $\mathrm{mRRA}=\frac{1}{N}\sum_{c=1}^{N}\mathrm{RRA}_c$, where the Relative Resistance Ability for corruption $c$ with respect to a baseline model reads as,
\begin{equation}
\mathrm{RRA}_c=\left(\frac{\sum_{s=1}^{3}\mathrm{NDS}_{c,s}}{\sum_{s=1}^{3}\mathrm{NDS}_{\mathrm{baseline},c,s}}-1\right),
\end{equation}
with $\mathrm{NDS}_{c,s}$ the nuScenes Detection Score under corruption $c$ at severity $s$.

\subsection{Uncertainty Modeling}

\PAR{Uncertainty Head.}
We augment the detector with an uncertainty head that outputs per-parameter variance estimates for each 3D bounding box.
Training follows the KL-divergence objective between the predicted parametric distribution and a Dirac distribution at the ground truth~\cite{kld}.
For the box center $(x,y,z)$, we model each coordinate with an independent Gaussian and regress the log-variance $u_i=\log\sigma_i^2$, yielding
\begin{equation}
    \mathcal{L}_{xyz} = \frac{1}{2}\sum_{i\in\{x,y,z\}}\left((\hat{x}_i - x_{i})^2 e^{-u_i} + u_i\right).
\end{equation}
To capture yaw while respecting angular periodicity, we use a von-Mises distribution.
With mean $\hat{\theta}$ and concentration $\kappa=e^{-u_\theta}$, the corresponding loss is
\begin{equation}
    \mathcal{L}_{\theta} = \log I_0(e^{-u_\theta}) + e^{-u_\theta}\bigl[1 - \cos(\hat{\theta}-\theta)\bigr],
\end{equation}
where $I_0$ is the modified Bessel function.
For numerical stability, we further include $\lambda_V\,\mathrm{ELU}(u_\theta - s_0)$~\cite{kld}.
Overall, this design remains deployment-friendly and yields uncertainty signals that are directly tied to the predicted box parameters, supporting downstream risk-aware decision-making and V2X data sharing.

\PAR{Uncertainty Calibration.}
To obtain calibrated confidence estimates, we apply deployment-friendly post-hoc calibration, following~\cite{beemelmanns2026q2u}.
For \emph{classification}, we calibrate logits $z$ with both Temperature Scaling (TS)~\cite{ts} and Platt Scaling (PS)~\cite{ps}: TS rescales logits as $p=\sigma(z/T)$, while PS fits an affine mapping $p=\sigma(a z + b)$ by minimizing the Negative Log-Likelihood on the calibration split, where $\sigma$ is the sigmoid function.
We evaluate classification calibration with the Detection Expected Calibration Error (D-ECE)~\cite{dece}, which measures how far predicted detection confidences deviate from empirical precision.
For \emph{regression} uncertainty, PS is not applicable, but instead we calibrate the predicted uncertainties with per-parameter temperatures with $\sigma_x\!\leftarrow\!\sigma_x/T_{\sigma_x}$, $\sigma_y\!\leftarrow\!\sigma_y/T_{\sigma_y}$, $\sigma_z\!\leftarrow\!\sigma_z/T_{\sigma_z}$, and $\kappa\!\leftarrow\!T_{\kappa}\kappa$.
All regression temperatures are fitted on a calibration set by minimizing the Miscalibration Area (MCA)~\cite{kuleshov2018accurate}, defined as the area between the nominal coverage curve and the empirically observed coverage of centered prediction intervals.
We report regression calibration as MCA$_{xyz}$ for the 3D centroid position and MCA$_{\theta}$ for orientation.
During calibration and evaluation, we apply a detection threshold of $\tau=0.3$ (discarding lower-confidence detections) and use a sequential $30\%$ split of the nuScenes val set for calibration and the remaining $70\%$ split for evaluation, for both regression and classification.

\subsection{Model Card and Data Card}
We create a Model Card documenting the CMT model architecture, training protocol, intended use, and known limitations, as well as a Data Card for the used dataset, covering data collection, annotation, biases, and class distributions.
Both documents are provided in the supplementary material.

\begin{table*}
\centering
\caption{
\textbf{Robustness of LiDAR-camera 3D Detectors.}
We report relative robustness score RRA, using BEVFusion~\cite{liu2022bevfusion} as baseline, for models trained with and without \emph{Masked-Modal Training}, together with aggregated mRRA and clean NDS.
}
\label{tab:1_robustness}
\resizebox{\textwidth}{!}{
\begin{tabular}{cc|cccccccccc|cc}
\toprule
\multirow{3}{*}{Model} & Masked-  & \multicolumn{10}{c|}{RRA$\uparrow$} & \multirow{3}{*}{mRRA$\uparrow$} & \multirow{3}{*}{NDS$\uparrow$} \\
                       & Modal    & Beams & Bright- & Dark- & Fog & Missing & Motion & Points & Snow & Spatial & Temp. &  &  \\
                       & Training & Red.  & ness    & ness  &     & Cam.    & Blur   & Red.   &      & Mis.    & Mis.   &  &  \\
\midrule
\multirow{2}{*}{DeepInteraction~\cite{yang2022deepinteraction}}
      &              & -6.361 & -3.150 & -7.215 & -25.037 & -16.386 & -7.077 & -2.188 & -5.149 & 0.212 & 0.145 & -7.221 & 0.691 \\
      & $\checkmark$ & -4.567 &  1.129 & 0.660  & 0.062   & 0.508   & -5.595 & -2.541 & -2.472 & -3.295 & 3.623 & -1.249 & 0.706 \\
     \cmidrule{2-14}
\multirow{2}{*}{TransFusion~\cite{bai2021pointdsc}}
      &              & -7.210 & 1.799 & 1.146 & -0.552 & 0.340 & -5.412 & -3.296 & -4.220 & -3.626 & 3.850 & -1.718 & 0.708 \\
      & $\checkmark$ & -7.313 & 1.476 & 0.973 & -0.546 & 0.714 & -4.242 & -4.537 & -5.311 & -3.984 & 3.240 & -1.953 & 0.705 \\
     \cmidrule{2-14}
\multirow{2}{*}{\shortstack{BEVFusion~\cite{liu2022bevfusion}\\(Baseline)}}
      &              &   0.000 &  0.000 &  0.000 &  0.000 &  0.000 &  0.000 &  0.000 &  0.000 &   0.000 & 0.000 &  0.000 & 0.714 \\
      & $\checkmark$ & -12.612 & -0.613 & -0.727 & -2.841 & -1.303 & -8.699 & -8.592 & -8.639 & -11.547 & 0.315 & -5.526 & 0.687 \\
     \cmidrule{2-14}
\multirow{2}{*}{IS-Fusion~\cite{yin2024isfusion}}
      &              & 1.663 & -0.096 & -1.252 & 6.079 & -3.138 & 0.604 & -0.241 & -5.468 & 3.679 & 5.731 & 0.756 & 0.725 \\
      & $\checkmark$ & 3.684 & 2.291 & 1.267 & 3.890 & 0.920 & \textbf{3.994} & 1.691 & -2.351 & 4.513 & 7.177 & 2.708 & \textbf{0.737} \\
     \cmidrule{2-14}
\multirow{2}{*}{SparseFusion~\cite{xie2023sparsefusion}}
       &              & 4.264 & \textbf{3.179} & \textbf{1.821} & 4.429 & 0.297 & 0.280 & 3.242 & 1.887 & 3.699 & 7.228 & 3.033 & 0.732 \\
       & $\checkmark$ & 13.709 & 3.044 & 1.508 & 5.701 & \textbf{3.795} & 0.474 & 4.762 & 2.785 & 6.101 & 8.096 & 4.997 & 0.733 \\
      \cmidrule{2-14}
\multirow{2}{*}{CMT~\cite{yan2023cross}}
       &              & 9.459 & -4.607 & -5.250 & 3.815 & -6.899 & -5.320 & 4.253 & 1.453 & 6.280 & 6.140 & 0.932 & 0.719 \\
       & \cb$\checkmark$ & \cb\textbf{18.642} & \cb-1.138 & \cb-0.096 & \cb\textbf{9.398} & \cb2.041 & \cb-0.841 & \cb\textbf{8.213} & \cb\textbf{9.887} & \cb\textbf{17.053} & \cb\textbf{8.448} & \cb\textbf{7.161} & \cb0.729 \\
\bottomrule
\end{tabular}
}
\end{table*}

\section{Experiments}
We evaluate the explainability faithfulness and uncertainty quality of our approach using the nuScenes dataset~\cite{caesar2020nuscenes} and MultiCorrupt~\cite{beemelmanns2024multicorrupt} for robustness evaluation.
Our approach shows strong robustness, generates faithful saliency maps, and precise uncertainty estimates, establishing a Trustworthy AI component within the AD stack.

\PAR{Experimental Setup.}
We train the modified detector on the nuScenes~\cite{caesar2020nuscenes} training set for 24 epochs following the original training protocol~\cite{yan2023cross} and evaluate its accuracy using the full validation split.
The model processes LiDAR sweeps and six cameras, encoded into a token set consisting of a $180\times180$ BEV LiDAR grid ($S_{pc}=32{,}400$ tokens) and $6$ camera feature maps of size $40\times100$ ($S_{img}=24{,}000$ tokens), yielding $S=56{,}400$ sensor tokens in total.
We extract cross-attention weights from the $L=6$ transformer decoder layers with $H=8$ heads.

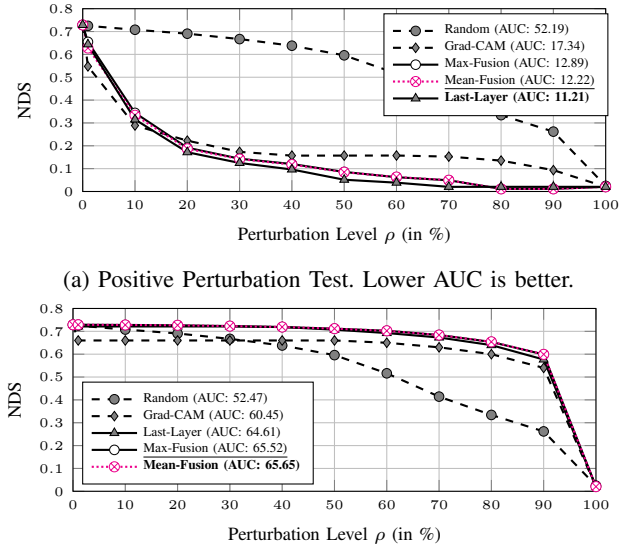
\begin{figure}[t]
\centering
\begin{subfigure}{0.475\textwidth}
\centering
\begin{tikzpicture}
\begin{axis}[
    width=8.5cm, height=4.0cm,
    xlabel={Perturbation Level $\rho$ (in \%)},
    xlabel near ticks,
    ylabel={NDS},
    ylabel near ticks,
    ylabel style={font=\scriptsize},
    xlabel style={font=\scriptsize},
    xticklabel style={font=\tiny},
    yticklabel style={font=\tiny},
    ymin=0, ymax=0.8,
    xmin=0, xmax=100,
    xtick={0,10,20,30,40,50,60,70,80,90,100},
    ytick={0.0,0.1,...,0.8},
    grid=both,
    legend style={font=\tiny, at={(0.575,0.975)}, anchor=north west, row sep=-0.09cm},
    legend cell align={left}
]
\pgfplotsset{
  maxstyle/.style={
    thick, solid, black,
    mark=*, mark size=1.8pt,
    every mark/.append style={solid, fill=white, draw=black, line width=0.2pt}
  },
  minstyle/.style={
    thick, dotted, black,
    mark=square*, mark size=1.8pt,
    every mark/.append style={solid, fill=gray, draw=black, line width=0.2pt}
  },
  meanstyle/.style={
    thick, densely dotted, magenta,
    mark=otimes*, mark size=1.9pt,
    every mark/.append style={solid, fill=white, draw=magenta, line width=0.2pt}
  },
  laststyle/.style={
    thick, solid, black,
    mark=triangle*, mark size=1.9pt,
    every mark/.append style={solid, fill=gray, draw=black, line width=0.2pt}
  },
  gradcamstyle/.style={
    thick, dashed, black,
    mark=diamond*, mark size=1.8pt,
    every mark/.append style={solid, fill=gray, draw=black, line width=0.2pt}
  },
  randomstyle/.style={
    thick, dashed, black,
    mark=*, mark size=1.8pt,
    every mark/.append style={solid, fill=gray, draw=black, line width=0.2pt}
  },
}

\addplot+[randomstyle] coordinates {
(0,0.7288) (1,0.7249) (10,0.7078) (20,0.6906) (30,0.6670)
(40,0.6380) (50,0.5959) (60,0.5163) (70,0.4143)
(80,0.3341) (90,0.2620) (100,0.0246)
};
\addlegendentry{Random (AUC: 52.19)}

\addplot+[gradcamstyle] coordinates {
(0,0.7288) (1,0.5475) (10,0.2888) (20,0.2224) (30,0.1734)
(40,0.1570) (50,0.1571) (60,0.1574) (70,0.1527)
(80,0.1352) (90,0.0934) (100,0.0204)
};
\addlegendentry{Grad-CAM (AUC: 17.34)}

\addplot+[maxstyle] coordinates {
(0,0.7288) (1,0.6547) (10,0.3440) (20,0.1916) (30,0.1435)
(40,0.1202) (50,0.0856) (60,0.0626) (70,0.0491)
(80,0.0115) (90,0.0115) (100,0.0204)
};
\addlegendentry{Max-Fusion (AUC: 12.89)}

\addplot+[meanstyle] coordinates {
(0,0.7288) (1,0.6279) (10,0.3336) (20,0.1868) (30,0.1419)
(40,0.1178) (50,0.0844) (60,0.0633) (70,0.0494)
(80,0.0115) (90,0.0115) (100,0.0204)
};
\addlegendentry{\underline{{Mean-Fusion (AUC: 12.22)}}}

\addplot+[laststyle] coordinates {
(0,0.7288) (1,0.6439) (10,0.3152) (20,0.1718) (30,0.1249)
(40,0.0963) (50,0.0515) (60,0.0386) (70,0.0203)
(80,0.0204) (90,0.0204) (100,0.0204)
};
\addlegendentry{\textbf{Last-Layer (AUC: 11.21)}}

\end{axis}
\end{tikzpicture}
\caption{Positive Perturbation Test. Lower AUC is better.}
\label{fig:positive_pert}
\end{subfigure}
\begin{subfigure}{0.475\textwidth}
\begin{tikzpicture}
\begin{axis}[
    width=8.5cm, height=4.0cm,
    xlabel={Perturbation Level $\rho$ (in \%)},
    xlabel near ticks,
    ylabel={NDS},
    ylabel near ticks,
    ylabel style={font=\scriptsize},
    xlabel style={font=\scriptsize},
    xticklabel style={font=\tiny},
    yticklabel style={font=\tiny},
    ymin=0, ymax=0.8,
    xmin=0, xmax=100,
    xtick={0,10,20,30,40,50,60,70,80,90,100},
    ytick={0.0,0.1,...,0.8},
    grid=both,
    legend style={font=\tiny, at={(0.0175,0.60)}, anchor=north west, row sep=-0.09cm},
    legend cell align={left}
]
\pgfplotsset{
  maxstyle/.style={
    thick, solid, black,
    mark=*, mark size=1.8pt,
    every mark/.append style={solid, fill=white, draw=black, line width=0.2pt}
  },
  minstyle/.style={
    thick, dotted, black,
    mark=square*, mark size=1.8pt,
    every mark/.append style={solid, fill=gray, draw=black, line width=0.2pt}
  },
  meanstyle/.style={
    thick, densely dotted, magenta,
    mark=otimes*, mark size=1.9pt,
    every mark/.append style={solid, fill=white, draw=magenta, line width=0.2pt}
  },
  laststyle/.style={
    thick, solid, black,
    mark=triangle*, mark size=1.9pt,
    every mark/.append style={solid, fill=gray, draw=black, line width=0.2pt}
  },
  gradcamstyle/.style={
    thick, dashed, black,
    mark=diamond*, mark size=1.8pt,
    every mark/.append style={solid, fill=gray, draw=black, line width=0.2pt}
  },
  randomstyle/.style={
    thick, dashed, black,
    mark=*, mark size=1.8pt,
    every mark/.append style={solid, fill=gray, draw=black, line width=0.2pt}
  },
}

\addplot+[randomstyle] coordinates {
(0,0.7288) (1,0.7249) (10,0.7078) (20,0.6906) (30,0.6670)
(40,0.6380) (50,0.5959) (60,0.5163) (70,0.4143)
(80,0.3341) (90,0.2620) (100,0.0246)
};
\addlegendentry{Random (AUC: 52.47)}

\addplot+[gradcamstyle] coordinates {
(0,0.7288) (1,0.66) (10,0.66) (20,0.66) (30,0.66)
(40,0.66) (50,0.66) (60,0.65) (70,0.63) (80,0.60)
(90,0.54) (100,0.02)
};
\addlegendentry{Grad-CAM (AUC: 60.45)}

\addplot+[laststyle] coordinates {
(0,0.7288) (1,0.7213) (10,0.7213) (20,0.7214) (30,0.7213)
(40,0.7187) (50,0.7073) (60,0.6925) (70,0.6730) (80,0.6396)
(90,0.5767) (100,0.0204)
};
\addlegendentry{Last-Layer (AUC: 64.61)}

\addplot+[maxstyle] coordinates {
(0,0.7288) (1,0.7289) (10,0.7279) (20,0.7264) (30,0.7230)
(40,0.7186) (50,0.7118) (60,0.7024) (70,0.6839) (80,0.6535)
(90,0.5988) (100,0.0204)
};
\addlegendentry{\underline{Max-Fusion (AUC: 65.52)}}

\addplot+[meanstyle] coordinates {
(0,0.7288) (1,0.7288) (10,0.7279) (20,0.7265) (30,0.7229)
(40,0.7189) (50,0.7118) (60,0.7024) (70,0.6838) (80,0.6541)
(90,0.5992) (100,0.0204)
};
\addlegendentry{\textbf{Mean-Fusion (AUC: 65.65)}}

\end{axis}
\end{tikzpicture}
\caption{Negative Perturbation Test. Higher AUC is better.}
\label{fig:negative_pert}
\end{subfigure}
\caption{
\textbf{XAI Faithfulness Test.}
NuScenes Detection Score (NDS) under increasing perturbation level $\rho$. The proposed \emph{Mean-Fusion} yields the best overall trade-off for both tests.
}
\label{fig:perturbation_test}
\end{figure}

\begin{figure}[ht]
\centering
\begin{subfigure}{0.475\textwidth}
\includegraphics[width=\linewidth]{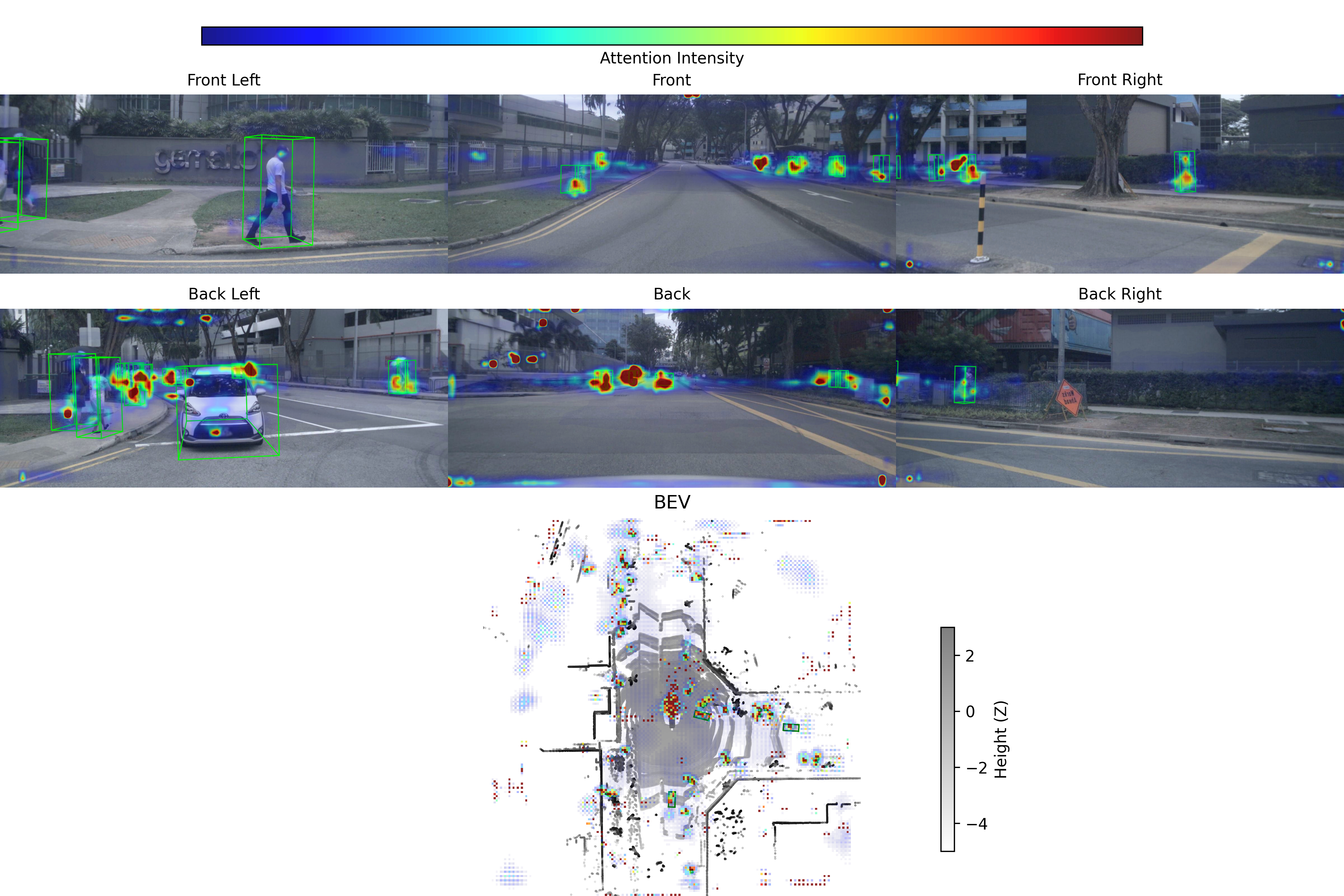}
\caption{
\textbf{Attention Maps.}
Visualization of the multi-modal saliency maps across multi-view images and LiDAR point cloud.
}
\label{fig:overview}
\end{subfigure}
\begin{subfigure}{0.475\textwidth}
\includegraphics[width=\linewidth]{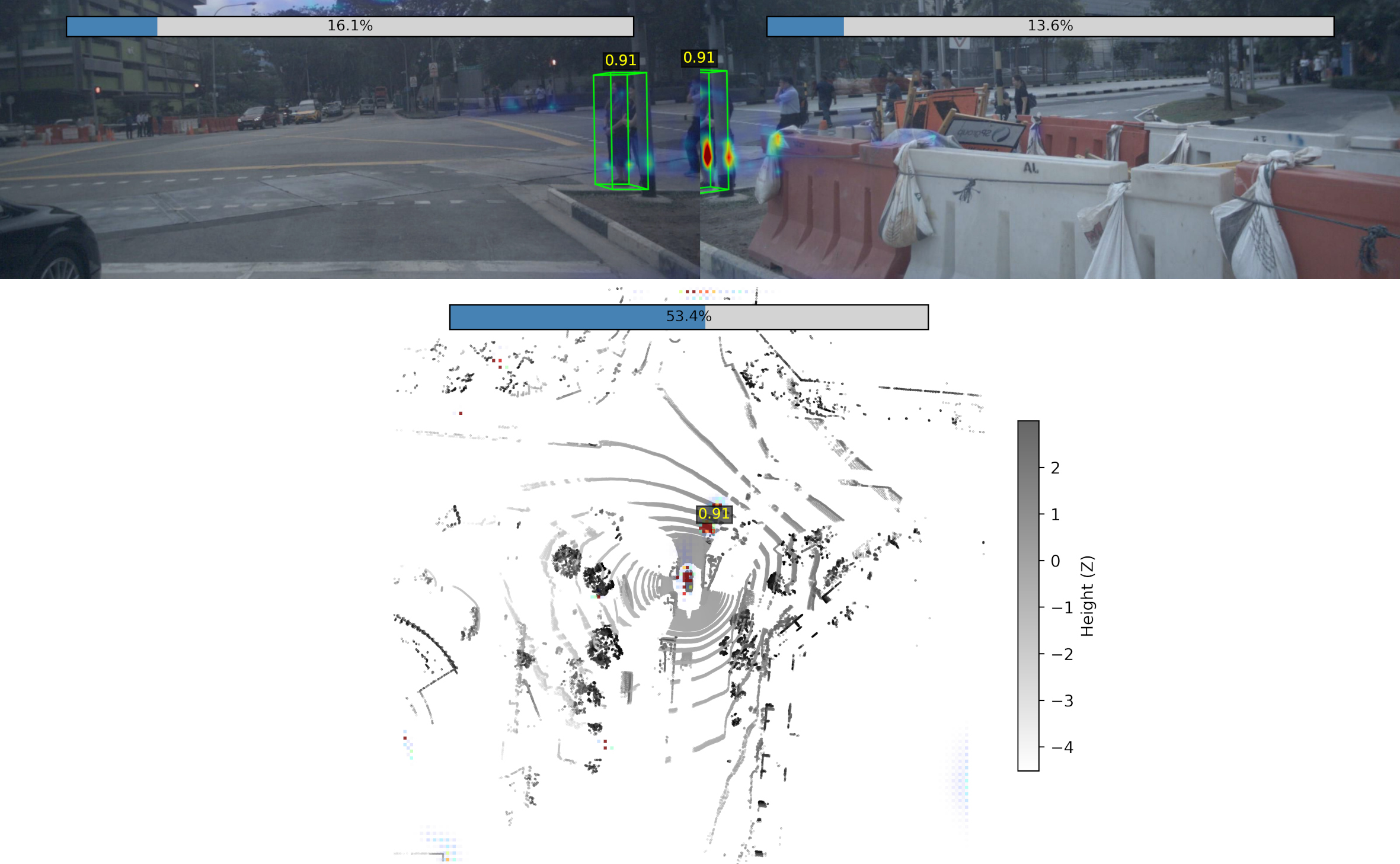}
\caption{
\textbf{Sensor Contribution.}
Contribution (in \%) for a detection shown for the two most contributing cameras and the point cloud. Remaining attention is shared among the other camera images.
}
\label{fig:contribution}
\end{subfigure}
\caption{
\textbf{XAI Attention Maps Visualization.}
}
\label{fig:xai_vis}
\end{figure}

\PAR{XAI Faithfulness.}
We evaluate explanation faithfulness via perturbation tests on LiDAR and camera inputs, and provide qualitative examples of the obtained saliency maps and modality contributions in Fig.~\ref{fig:xai_vis}.
For LiDAR, the point cloud is discretized into a $180\times 180$ BEV grid; given a masking fraction $\rho$, we remove all LiDAR points within selected grid cells.
For camera images, we mask a fraction $\rho$ of pixels per view by replacing them with the image's mean color to suppress information.
We apply these perturbations guided by our \emph{Mean-Fusion} approach for \emph{Positive Perturbation} (masking high-attention regions) and \emph{Negative Perturbation} (masking low-attention regions).
For positive perturbations, a larger score drop at the same $\rho$ indicates that the saliency maps capture regions that are critical for detection, while a stable detection score under negative perturbations suggests that low-attention regions are indeed less relevant.
To summarize this behavior, we report the area under the NDS-$\rho$ curve (AUC): for positive perturbations, a \emph{lower} AUC indicates more faithful localization of critical regions (faster degradation), whereas for negative perturbations, a \emph{higher} AUC indicates better robustness to masking irrelevant regions.
We compare our approach against four baselines: (i) \emph{Grad-CAM}~\cite{Selvaraju2017GradCAM}, (ii) \emph{Max-Fusion}, which uses max-pooling across all decoder layers; (iii) \emph{Last-Layer}, utilizing only the final decoder layer, and (iv) \emph{Random}, where a randomized fraction $\rho$ of LiDAR grid cells and image pixels are masked.
We also considered D-RISE~\cite{Petsiuk2021drise} as baseline; however, scaling its random masking procedure to multi-view camera and LiDAR tokens causes infeasible runtimes, and was therefore omitted.
Across perturbation levels, our \emph{Mean-Fusion} approach outperforms random masking and exceeds the sensitivity of Grad-CAM~\cite{Selvaraju2017GradCAM}, as shown in Fig.~\ref{fig:perturbation_test}.

\PAR{Robustness Assessment.}
We evaluate robustness on MultiCorrupt~\cite{beemelmanns2024multicorrupt} using ten LiDAR-camera corruptions at three severity levels.
Table~\ref{tab:1_robustness} reports corruption-specific RRA and aggregated mRRA, together with the nuScenes Detection Score (NDS) on the clean validation split.
Each multi-modal detector is trained both with and without masked-modal training to isolate its effect on robustness.
Masked-modal training improves our detector's robustness leading to the most robust model among other approaches.
This indicates that modality dropout during training reduces brittle cross-modal dependencies, yielding a perception module that remains reliable and trustworthy under challenging conditions.

\PAR{Uncertainty Calibration Quality.}
We assess how well the detector’s confidence and predicted box uncertainties are calibrated, since miscalibration undermines trustworthiness even when raw accuracy is high.
For \emph{classification}, the detection scores are calibrated with TS and PS, where PS yields the best calibration performance in terms of D-ECE, as shown in Table~\ref{tab:uncertainty_calibration}.
For \emph{regression}, we apply per-parameter TS to the predicted variances, yielding lower MCA for both centroid position and orientation.
\begin{table}[h]
\caption{
\textbf{Calibration Quality.}
We report classification calibration error (D-ECE), regression calibration accuracy for bounding box centroid (MCA$_{xyz}$) and orientation (MCA$_{\theta}$).
}
\centering
\begin{tabular}{cccccc}
\toprule
\multirow{2}{*}{Method} & \multicolumn{1}{c}{Cls.} & \multicolumn{2}{c}{Regression} & \multicolumn{2}{c}{Accuracy} \\
\cmidrule(lr){2-2}\cmidrule(lr){3-4}\cmidrule(lr){5-6}
                        & D-ECE$\downarrow$         & MCA$_{xyz}\!\downarrow$      & MCA$_{\theta}\!\downarrow$       & mAP$\uparrow$               & NDS$\uparrow$ \\
\midrule
Uncal.       &           11.601  &            1.187  &             7.322 & 0.703 & 0.729 \\
TS~\cite{ts} &            1.597  & \cb\textbf{0.781} & \cb\textbf{6.734} & 0.703 & 0.729 \\
PS~\cite{ps} & \cb\textbf{1.018} &               n/a &               n/a & 0.703 & 0.729 \\
\bottomrule
\end{tabular}
\label{tab:uncertainty_calibration}
\end{table}

\section{Prototype}
We deploy a prototype of the Trustworthy AI perception component on \emph{karl.}~\cite{karl}, an automated driving research platform, demonstrating the real-world integration of explainability, uncertainty quantification, and calibration in a live AD stack.
For simplicity and to reduce synchronization and calibration overhead under limited training data, we disable the camera branch and focus on LiDAR-only detection.
We use a reduced BEV LiDAR grid of $128\times128$ and keep the remaining transformer detector identical to CMT~\cite{yan2023cross}, i.e., a decoder with $L=6$ transformer layers and $H=8$ attention heads, extended with the uncertainty prediction head.

\begin{figure}[t]
\centering
\begin{subfigure}{0.475\textwidth}
\includegraphics[clip, trim=3.75cm 12cm 0cm 14cm, width=\linewidth]{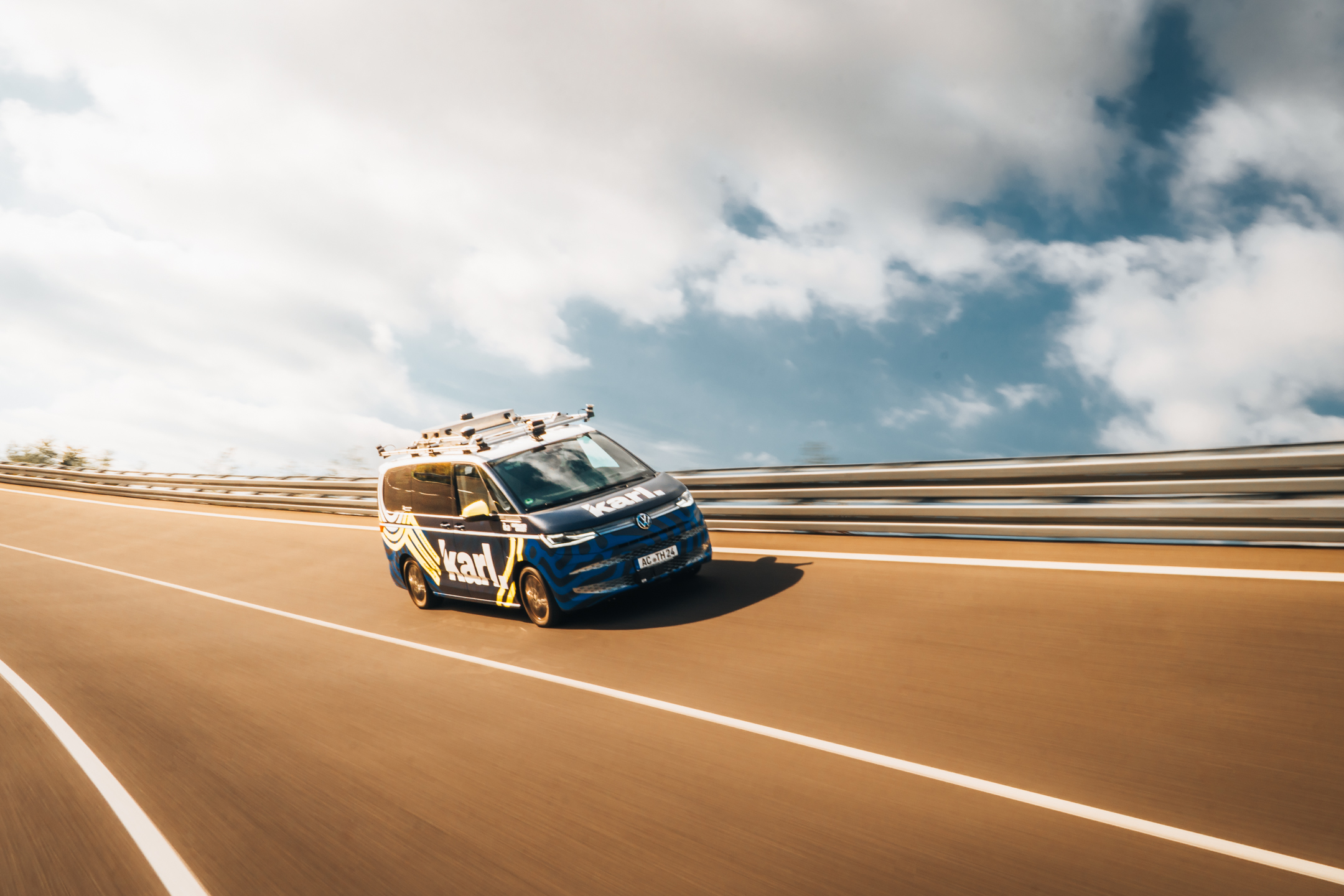}
\caption{
\textbf{karl.}
Research vehicle used to deploy the proposed approach.
}
\label{fig:karl}
\end{subfigure}
\begin{subfigure}{0.475\textwidth}
\includegraphics[clip, trim=0cm 8.00cm 2cm 3.05cm, width=\linewidth]{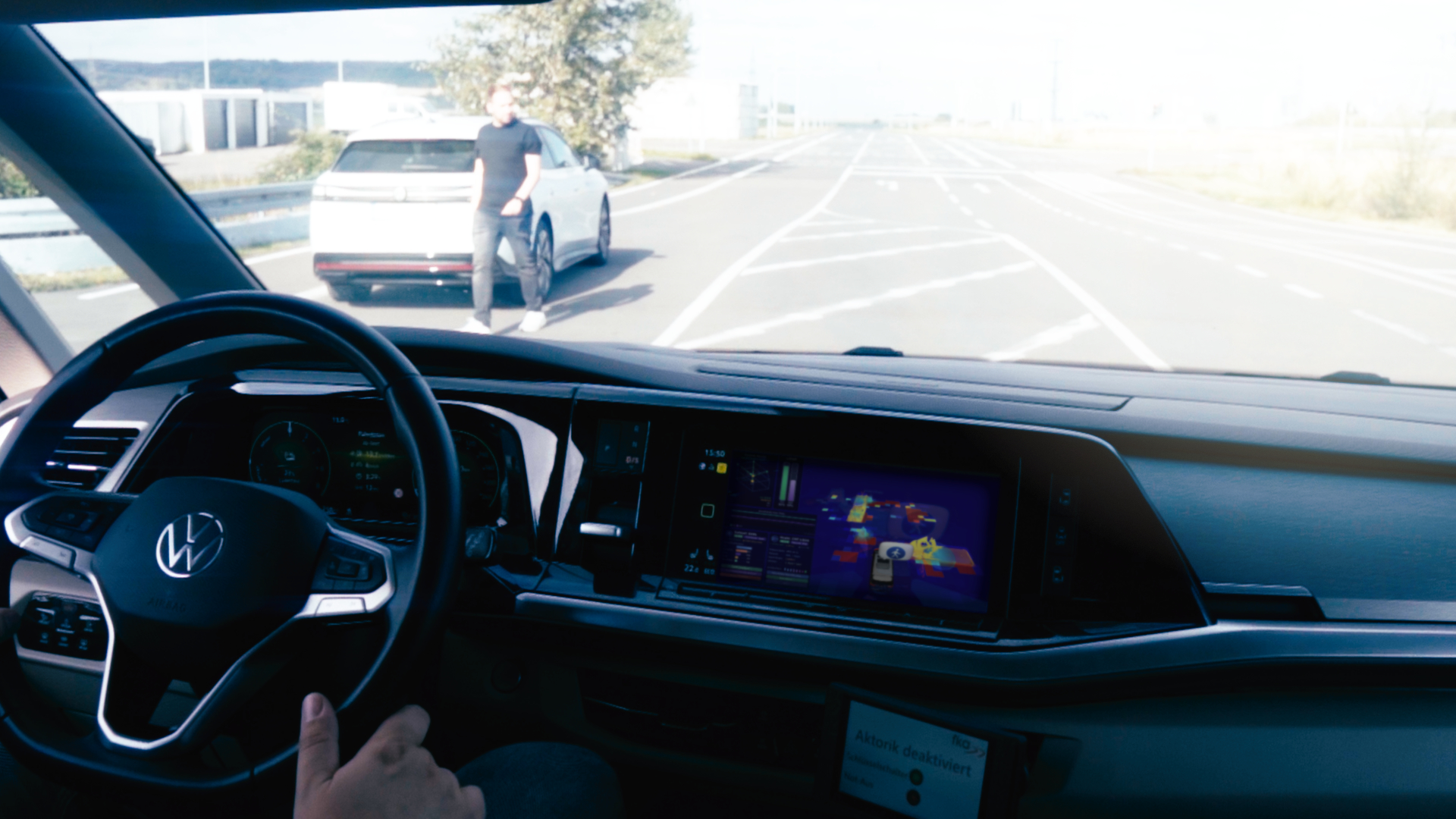}
\caption{
\textbf{XAI Interface.}
Embedded into the vehicle's dashboard.
}
\label{fig:xai_interface_integration}
\end{subfigure}
\begin{subfigure}{0.475\textwidth}
\includegraphics[width=\linewidth]{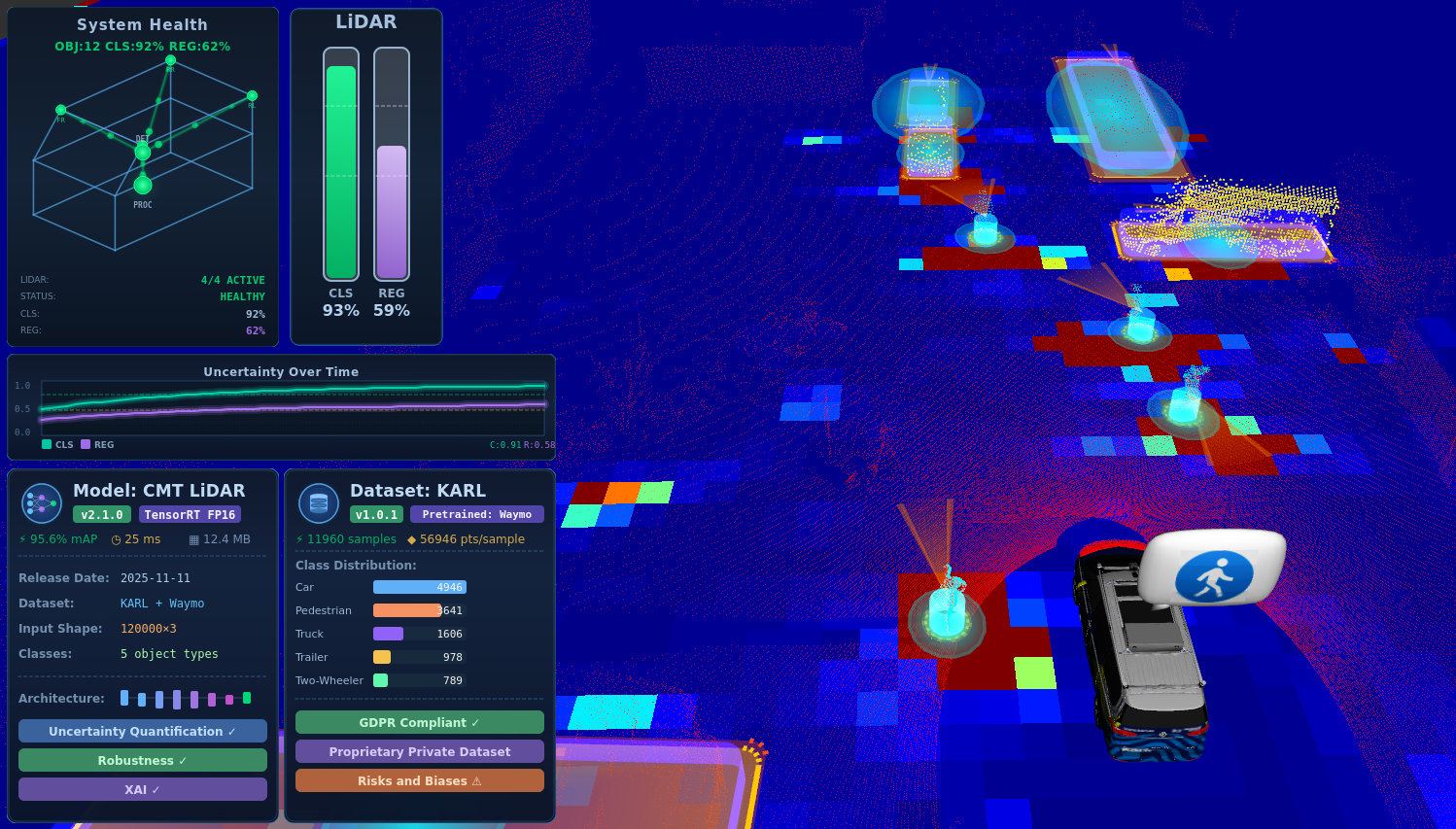}
\caption{
\textbf{XAI Interface in RViz.}
Visualizations showcasing system health, dynamic uncertainty states, Model Cards and Data Cards for documentation, and attention-based saliency maps overlaid on the 3D scene in real time.
Best viewed zoomed in.
}
\label{fig:xai_interface}
\end{subfigure}
\caption{
\textbf{Vehicle Prototype Integration.}
}
\label{fig:prototype}
\end{figure}

The model is finetuned and calibrated using an in-house collected and annotated dataset.
For deployment, we accelerate the detection module with TensorRT and FP16 quantization, and implement the detection pipeline as a ROS~2~Jazzy node.
The real-time XAI Interface is implemented as an RViz plugin, enabling the visualization of Model Cards and Data Cards, model detections including their uncertainty estimates, and the attention-based saliency maps, as shown in Figure~\ref{fig:prototype}.
In our vehicle setup, we fuse two front-facing Ouster OS1-128 LiDARs as input to the transformer-based detector and achieve an average inference latency of $24.4\,\mathrm{ms}$ with a 99th-percentile tail latency of $28.8\,\mathrm{ms}$ using an NVIDIA RTX~4090.

\section{Conclusion}
We presented a Trustworthy AI perception component for autonomous driving that combines attention-based explainability, uncertainty quantification with calibration, and robustness assessment.
The key premise is that Trustworthy AI requires both \emph{interpretability} and \emph{reliability}: explanations support debugging and auditability, calibrated uncertainties quantify predictive reliability, and robust architectural choice and training improve stability under distribution shifts and sensor degradations.
Our perturbation-based evaluation provides evidence that attention yields more faithful saliency maps than baseline methods.
Calibration further improves the reliability of both classification and regression uncertainties without affecting accuracy.
Finally, we demonstrated a prototype integration with real-time visualization of detections, attention maps, and uncertainty estimates, indicating practical feasibility of the approach.
Since the vehicle prototype is currently LiDAR-only and does not yet validate our multi-modal robustness claims, future work will extend it to full LiDAR-camera fusion and evaluate an LLM integration for a more interactive XAI Interface.

\bibliographystyle{IEEEtran}
\bibliography{root}

\clearpage

\end{document}